\documentclass{article}

\usepackage{spconf,amsmath,graphicx}
\usepackage{subfigure}

\usepackage{tikz}
\usetikzlibrary{spy,backgrounds}
\pgfdeclarelayer{background}
\pgfdeclarelayer{foreground}
\pgfsetlayers{background,main,foreground}
\title{Off-the-grid model based deep learning (O-MoDL)}

\name{Aniket Pramanik, Hemant Aggarwal, Mathews Jacob\thanks{This work is supported by NIH 1R01EB019961-01A1.}}

\address{University of Iowa, Iowa, USA}

\begin{document}

%\ninept

%

\maketitle

\begin{abstract}
We introduce a model based off-the grid image reconstruction algorithm using deep learned priors. The main difference of the proposed scheme with current deep learning strategies is the learning of non-linear annihilation relations in Fourier space. We rely on a model based framework, which allows us to use a significantly smaller deep network, compared to direct approaches that also learn how to invert the forward model. Preliminary comparisons against image domain MoDL approach demonstrates the potential of the off-the-grid formulation. The main benefit of the proposed scheme compared to structured low-rank methods is the quite significant reduction in computational complexity. 
\end{abstract}

\begin{keywords}
off-the-grid, CNN, MRI
\end{keywords}

\section{Introduction}
\label{sec:intro} 
The recovery of images from undersampled Fourier measurements is a classic problem in MRI and several other modalities. The popular approach is to constrain the reconstructions using compactness priors including sparsity. Several researchers have recently introduced off-the-grid continuous domain priors that are robust to discretization errors \cite{gregSIAM2016,haldar2014low}, which provide significantly improved image quality in a range of applications. However, the main challenge is the significant increase in  computational complexity. 

Recently, several researchers have introduced deep learning methods as fast and efficient alternatives to compressed sensing algorithms. Current approaches can be categorized into direct and model based strategies. The direct approaches directly estimate the images from the undersampled measurements or their transforms/features \cite{ye2018deep, unser2017}. These methods learn to invert the forward operator over the space/manifold of images. While this approach is more popular, a challenge with these schemes is the need to learn the inverse, which often requires large models (e.g. UNET); this often translates to the need for extensive training data. Model based strategies instead formulate the recovery as a penalized optimization problem using a forward model that mimics the acquisition scheme \cite{aggarwal2017modl}; the penalty is designed to encourage the reconstructed image to be close to the space/manifold of images. Since they rely on an explicit forward model, they often only require a much smaller network than direct methods, which significantly reduces the training data demand. Recent studies also show the benefit in sharing the network parameters across iterations, end-to-end training, as well as using optimization blocks within the network \cite{aggarwal2017modl}.

The main focus of this work is to extend deep learning to the off-the-grid setting, similar to the structured low-rank setting in \cite{gregSIAM2016,haldar2014low}. Specifically, we consider an iterative reweighted least-squares (IRLS) formulation of our off-the-grid algorithm, termed as GIRAF. We show that the structure of GIRAF, which alternates between data consistency enforcement and denoising of the Fourier coefficients of the partial derivatives by projecting it to the constraint set, has extensive similarities to the MoDL framework. The main difference with MoDL is that the learning is performed in the Fourier domain, and the denoising network is linear; GIRAF relies on a residual convolution and deconvolution (flipped convolution) denoising filter-bank with shared filters, which is learned from the available k-space measurements in an iterative fashion. The residual filterbank projects the Fourier coefficients to the signal subspace, thus denoising them. To significantly reduce the computational complexity of these off-the-grid methods, we propose to use a deep non-linear convolutional neural network as the denoiser for the Fourier coefficients of the partial derivatives. Similar to GIRAF, we propose to use convolutional and deconvolutional blocks with shared parameters. The parameters of the deep network are learned in an end-to-end fashion as in MoDL. This work has connections with k-space deep learning approach \cite{han2018k}, which follows a direct approach of estimating images from measurements, bypassing a forward model. By contrast, we follow a model based strategy, which has the benefits discussed above. %\vspace{-1em}

We determine the utility of the proposed scheme in the single coil image recovery setting. Thanks to the MoDL formulation, few training datasets were sufficient to reliably learn the parameters of the model. The comparisons of the proposed scheme against MoDL shows the potential of the proposed scheme. Our future work will focus on the extension of the framework to the multi-coil setting and more elaborate comparisons with state of the art algorithms. 

\vspace{-2em}
\section{Problem Setting and Background}
\label{sec:algo}
We consider the recovery of the Fourier coefficients of a function $f(\mathbf r)$ from its noisy undersampled Fourier measurements $\mathbf b=\mathbf A \hat{\mathbf f}+\boldsymbol\eta$, where $\hat f$ denotes the Fourier transform of $f(\mathbf r)$, and $\mathbf A$ is an undersampling matrix, while $\eta$ is additive noise.

\subsection{Piecewise smooth images \& annihilation relations}
Recent work off-the-grid methods \cite{gregSIAM2016,haldar2014low} model the signal $f$ to be piecewise smooth, when the partial derivatives of $f$ vanish everywhere except on an edge set. Similar to \cite{gregSIAM2016}, we model the edge set as the zero-set of a bandlimited function $\mu$. With these assumptions, the first order partial derivatives of the signal satisfies $\nabla f.\mu=0$, which translates to $\widehat{\nabla f} \ast h = 0$, where $h\stackrel{\mathcal F}{\leftrightarrow} \mu$ is the bandlimited Fourier transform of $\mu$. The theory in  \cite{gregSIAM2016} shows that if the assumed size of the filter $h$ is greater than the true bandwidth, there exist several linearly independent finite impulse response (FIR) filters that satisfy $\widehat{\nabla f} \ast h_i = 0; i = 1,.., N$. The above relation can be expressed in a matrix form as 
\begin{equation}
\underbrace{\left[\begin{array}{c|c|c|c}
\mathcal G(\mathbf h_1)&\mathcal G(\mathbf h_2)&\ldots & \mathcal G(\mathbf h_N)
\end{array}\right]}_{\mathcal T(\mathbf H)} \widehat{\nabla f} = 0.
\end{equation}
Here, $\mathcal G(\mathbf h_i)$ is a block Toeplitz matrix that corresponds to 2-D convolution. i.e, $\mathcal G(\mathbf h_i)\widehat{\nabla f} = h_i \ast \widehat{\nabla f}$. Using commutativity of convolutions, the above relation can also be expressed as
\begin{equation}
\label{filterest}
\underbrace{\left[\begin{array}{c}
\mathcal G\left(\widehat{\mathbf f_x}\right)\\
\mathcal G\left(\widehat{\mathbf f_y}\right)
\end{array}\right]}_{\mathcal T(\widehat{\nabla f})} \mathbf H = 0.
\end{equation}
where $\mathcal G(\widehat{\mathbf f_x})$ is the block Toeplitz matrix constructed from $\widehat{\mathbf f_x}$.

\begin{figure}[t!]
\centering
\subfigure[Iterative algorithm]{\includegraphics[width=6.8cm,trim={0 6cm 0 5cm},clip]{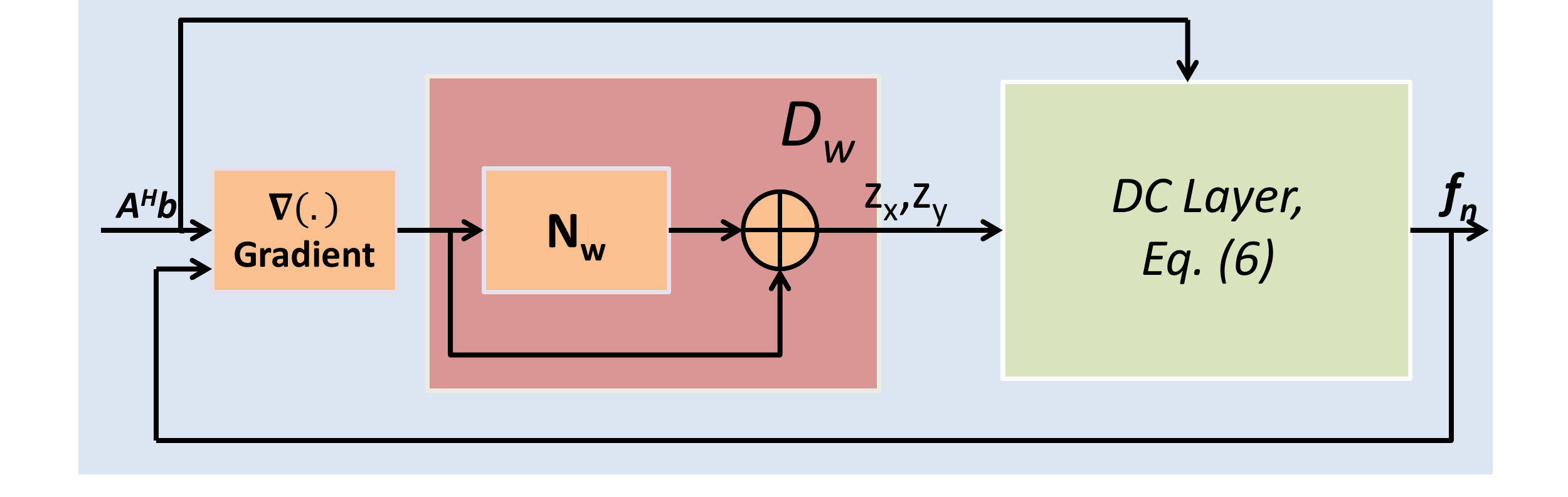}}
\subfigure[GIRAF residual denoising block]{\includegraphics[width=6.5cm,trim={0 8cm 0 6cm},clip]{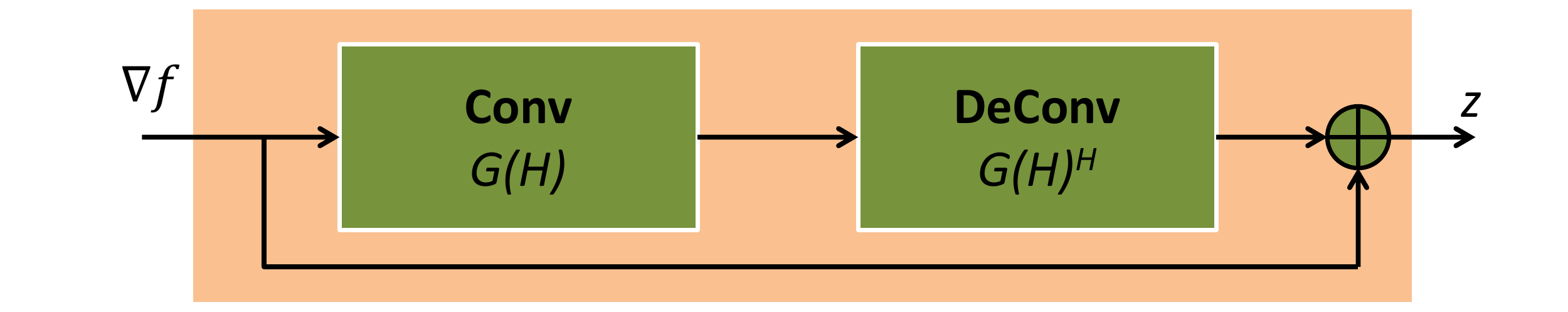}}
\subfigure[O-MoDL residual denoising block]{\includegraphics[width=8.7cm,trim={0 5cm 0 2.5cm},clip]{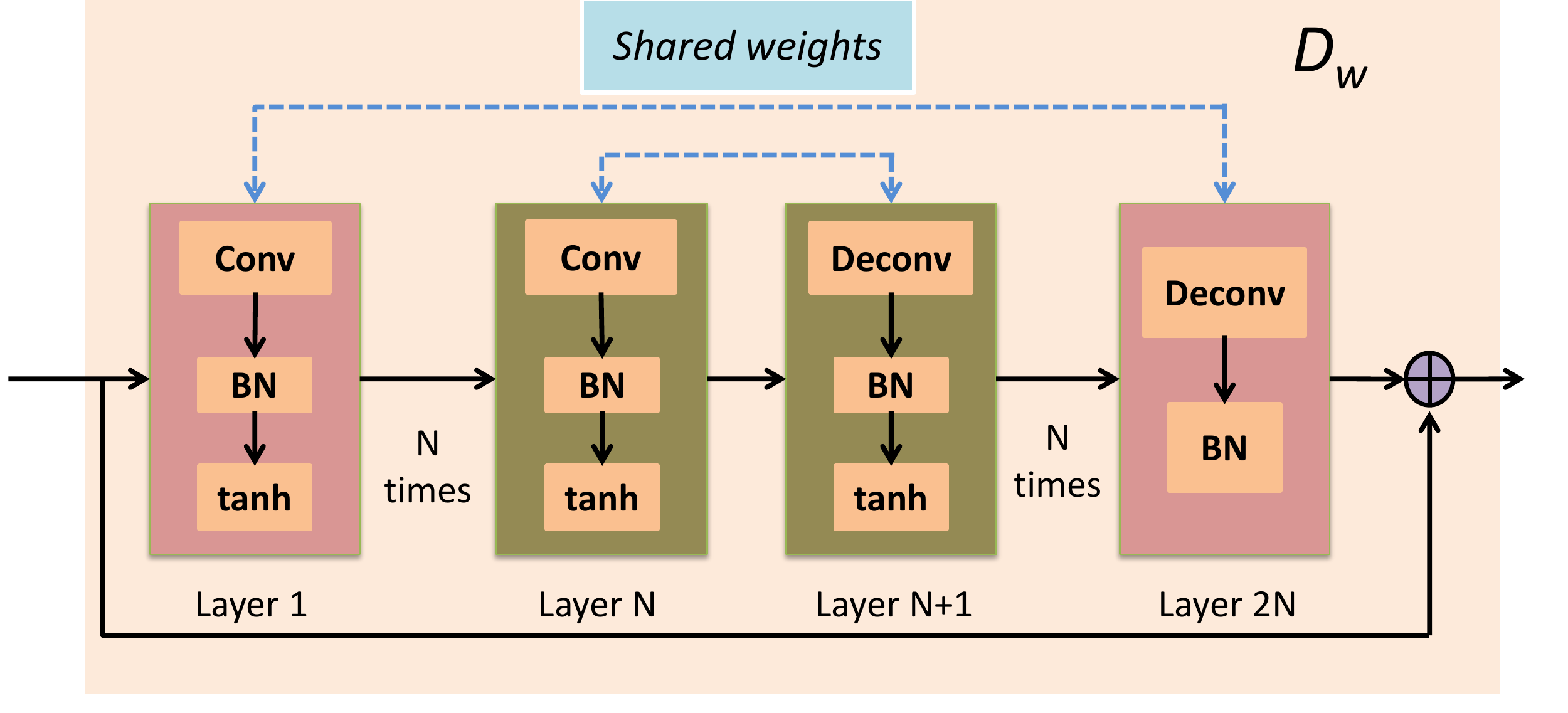}}\vspace{-1.5em}
%\subfigure[Unrolled deep network]{\includegraphics[width=8.7cm,trim={0 7cm 0 6cm},clip]{Unrolled.pdf}}
\label{architecture}
\caption{Outline of the recursive learning architectures. (a) Proposed iterative algorithm, which alternates between a residual denoiser block and data-consistency block. GIRAF and O-MoDL frameworks differs only in the structure of the residual denoiser blocks, shown in (b) and (c), respectively. (b). GIRAF residual denoising block, where the filters $\mathbf H$ are learned using structured low-rank matrix completion. (c) O-MoDL residual denoising block: the linear denoiser in GIRAF is replaced by a deep CNN architecture, which uses convolution and deconvolution blocks. The filter parameters of the corresponding convolution and deconvolution blocks are shared. %(d) The deep architecture obtained by unrolling the recursive structure in (a).
} 
\vspace{-1.2em}
\end{figure}

\subsection{GIRAF algorithm for off-the-grid image recovery}
In practice, the filters $h_i; i=1,..,N$ are unknown. Two step methods \cite{gregSIAM2016, haldar2014low} estimate the filters $h_i; i=1,..,N$ or equivalently $\mathcal T(\mathbf H)$ from fully sampled regions of Fourier space, which is also called as calibration regions, using \eqref{filterest}. When the Fourier samples are non-uniformly sampled, the above two step approach is not feasible. Note that \eqref{filterest} implies that the matrix $\mathcal T(\mathbf H)$ is low-rank. The unconstrained version of GIRAF algorithm poses the recovery as \begin{eqnarray}
\label{constrained}
\hat f = \arg \min_{\hat f} \lambda \left \|\mathcal T(\widehat{\nabla f}) \right\|_* + \|\mathcal A \hat f - \mathbf b\|^2 ,
\end{eqnarray} 
where $\|\cdot\|_*$ is the nuclear norm that encourages the matrix to be low-rank. GIRAF relies on an iterative least squares optimization problem, which majorizes the nuclear norm as $\|\mathbf{T} (\widehat{\nabla f})\|_*\leq  \| \mathbf{T} (\widehat{\nabla f}) \mathbf H\|_F^2$. This algorithm alternates between the estimation of $\mathbf H$ as $\mathbf H = \big[
\mathbf{T}^*(\mathbf{T}(\widehat{\nabla f})) + \epsilon 
 \mathbf I\big] ^{-1/4}$ and updating $f$ by
\begin{eqnarray}
\label{constrained}
\hat f = \arg \min_{\hat f} \lambda \|\mathcal T(\mathbf H)~\widehat{\nabla f} \|_F^2 + \|\mathcal A \hat f - \mathbf b\|^2
\end{eqnarray} 
A challenge with the alternating minimization strategy is the high computational complexity of the algorithm. Motivated by the fast computation offered by our MoDL framework \cite{aggarwal2017modl}, we introduce a deep learning solution.

%\begin{eqnarray}
%\nonumber
%\nabla f.\mu=0\\
%\nonumber
%or\hspace{4pt} equivalently,\hspace{4pt} \hat{f}_{x}\ast h_{i}=0,\\
%\hat{f}_{y}\ast h_{i}=0, ~ \forall i=1,.., N\\
%\nonumber
%where \begin{bmatrix}
%\hat{f}_{x} \\
%\hat{f}_{y}
%\end{bmatrix}
%=
%\begin{bmatrix}
%-j\omega_{x} \\
%-j\omega_{y}
%\end{bmatrix}
%\hat{f}=\widehat{\nabla f}
%\end{eqnarray}
%$\hat{f}_{x}$, $\hat{f}_{y}$ are the x and y gradients of $\hat{f}$ respectively.

\vspace{-2em}
\section{Off-the-grid model based deep learning (O-MoDL)}
We first show that the GIRAF algorithm can be viewed as a recursive network, which has alternating blocks imposing data consistency and denoising of the data. 

\vspace{-2em}
\subsection{Network structure of GIRAF }
We consider a penalized version of \eqref{constrained}, using an auxiliary variable $\mathbf z$:
\begin{eqnarray}\nonumber
\mathcal C\left\{\mathbf f, \mathbf z\right\} &=& \arg \min_{\mathbf f, \mathbf z} \|\mathcal A \hat f - \mathbf b\|^2+ \beta \|\widehat{\nabla f} - \mathbf z\|^2 +\\
&& \lambda \|\mathcal T(\mathbf H)~ \mathbf z \|_F^2 
%&& \mbox{subject  to    }\mathbf A \mathbf f=\mathbf b.
\end{eqnarray}
Note that the above formulation is equivalent to \eqref{constrained} when $\beta \rightarrow \infty$. We use an alternating algorithm to solve for $\mathbf f$ and $\mathbf z$:
\begin{eqnarray}\label{analytic}
\mathbf f_{n+1} &=& \arg \min_{\hat f}  \|\mathcal A \hat f - \mathbf b\|^2 + \beta \|\widehat{\nabla f} - \mathbf z_{n}\|^2 \\\label{smooth}
\mathbf z_{n+1} &=& \arg \min_{\mathbf z} \beta \|\nabla \widehat{f_{n+1}} - \mathbf z\|^2+ \lambda \|\mathcal T(\mathbf H)~\mathbf z\|_F^2
\end{eqnarray}
Problem \eqref{analytic} is analytically solved, when $\mathbf A$ is a sampling operator.
%\begin{equation}
%\label{analytic}
%\hat f[\mathbf m] = \left\{\begin{array}{ccc}
%\frac{ j(\omega_{x} z_{x}[\mathbf m]+\omega_{y}z_{y}[\mathbf m])}{ (\omega_{x}^{2}+\omega_{y}^{2})} & \mbox{if } & \mathbf m \mbox{ is sampled }\\
%b[m] & \mbox{else}
%\end{array}
%\%right. 
% \hspace{4pt} subject \hspace{4pt} to \hspace{4pt}\mathcal A (\hat f)=b
%\end{equation}
Solving \eqref{smooth}, we get $ \mathbf z = \left[\mathbf I ~+~\frac{\lambda}{\beta}~\mathcal T(\mathbf H)^H \mathcal T(\mathbf H) \right]^{-1}\widehat{\nabla f}$. Using matrix inversion lemma and assuming $\lambda <<\beta$, we approximate the solution as: 
\begin{equation}
\label{denoiser}
\mathbf z \approx \left[\mathbf I ~-~\frac{\lambda}{\beta}~\mathcal T(\mathbf H)^H \mathcal T(\mathbf H) \right]\widehat{\nabla f}.
\end{equation}
%--------FIGURE 1 --------------
\begin{figure}[t!]
\includegraphics[width=.5\textwidth]{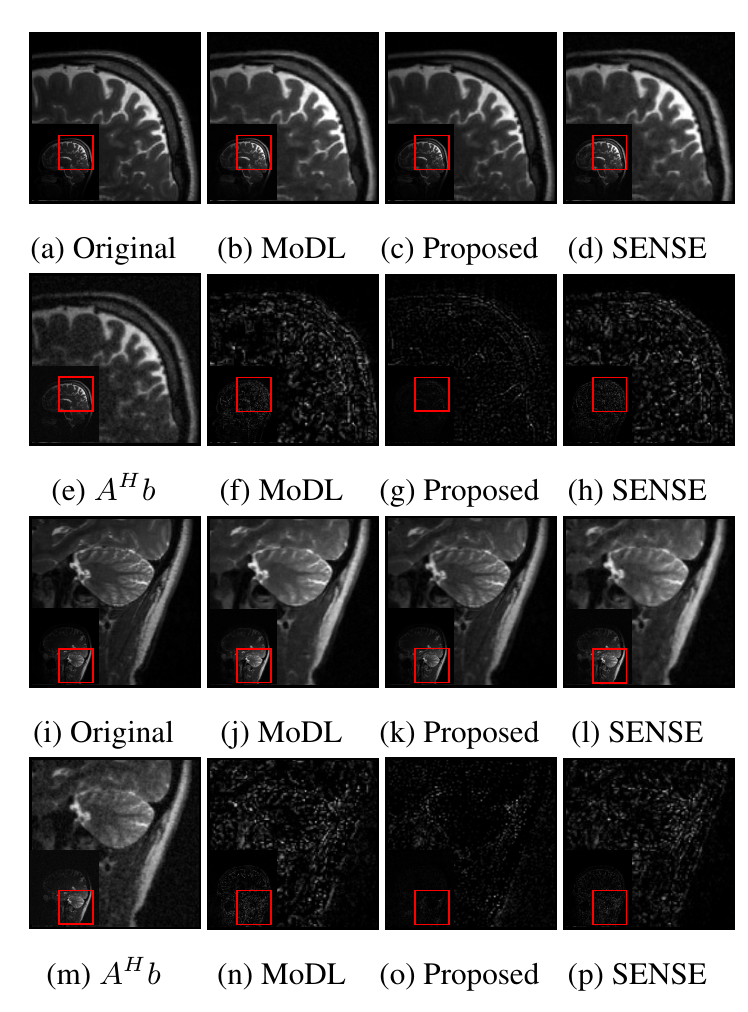}\vspace{-1.2em}
\caption{Reconstruction results of two-fold accelerated data. Figures (a)-(h) correspond to Coil 1 images while (i)-(p) correspond to Coil 6 images. (b) MoDL:16.33 db (c) O-MoDL: 18.57 db, (d) SENSE: 12.72 db, (f)-(h) are the error images. (j) MoDL,16.74 db (k) O-MoDL,18.98 db, (l) SENSE, 12.88 db, (n)-(p) are the error images.}\vspace{-1.2em}  
\end{figure}
%--------END FIGURE 1 --------------

Note from \eqref{filterest} that filters $\mathbf H$ are surrogates for the null space of $\mathcal T(\widehat{\nabla f})$. Hence, $\mathcal T(\mathbf H)^H\mathcal T(\mathbf H)\; \mathbf z$ can be viewed as the projection of $\mathbf z$ onto the null-space of $\mathbf H$; \eqref{denoiser} can be viewed as a residual block, which removes the null-space components of $\widehat{\nabla f}$. The alternating algorithm specified by \eqref{denoiser} and \eqref{analytic} can be unrolled to a deep architecture as in \cite{aggarwal2017modl}. 

\subsection{Deep learning in k-space}
\label{sec:model}
Motivated by the success of MoDL \cite{aggarwal2017modl}, we introduce a novel deep learning solution to reduce the computational complexity of GIRAF.  Specifically, we replace the denoiser in GIRAF, specified by \eqref{denoiser}, by a CNN, whose parameters are learned from exemplar data. The structure of O-MoDL is similar to GIRAF, involving two alternating blocks, $\mathcal D_{w} = \mathcal I-\mathcal N_w$ (denoiser) and DC (data consistency). We rely on an $n$-layer  O-MoDL $\mathcal N_w$ block consists of '$n$' convolution layers followed by '$n$' deconvolution (convolution with flipped filters) layers, followed by batch normalization (BN) and non-linear activation function. We chose the non-linear activation function as a hyperbolic tangent (tanh) activation function. Each convolution layer only retains the valid convolutions to eliminate boundary effects. The use of deconvolution blocks ensures the smaller image obtained after convolution grows back to its original size. The DC block refers to \eqref{analytic}. The main difference of this framework with GIRAF is the depth of the denoiser network (see Fig. \ref{architecture})  and the presence of non-linearities and batch normalization steps within the network. The key difference of the proposed scheme with MoDL is that the learning is performed in k-space, unlike most deep learning image recovery strategies. The denoiser network uses convolution and deconvolution layers similar to GIRAF, which share weights. This is a another distinction with MoDL, which uses a series of convolutional blocks with no weight sharing. Similar to MoDL, we also share the same network across iterations. 

%The training stage learns the parameters of the complex filters within $\mathcal N_w$, so as to annihilate the kspace coefficients of the gradients of all training images, while preserving noise; it behaves a projection to the null-space of $\mathcal T(\widehat{\nabla f})$. Since $\mathcal N_w$ is used within the residual block, the $D_{w}$ block is expected to work like a denoiser. Specifically, the noise estimated by $\mathcal N_w$ is subtracted from the kspace coefficients of the gradients, thus denoising the k-space data. 

\section{Results}
\label{sec:result}
%\subsection{MRI phantoms}
% In order to validate the proposed method, we perform Super-resolution recovery of piecewise constant MRI phantoms of dimensions 256 x 256, keeping $D_{w}$ block linear. We extract the low pass fourier samples within a square window of 64 x 64 to learn the annihilating filters described in section \ref{sec:model}. To recover image, we train the $D_{w}$ block with a dataset of 50 noisy undersampled kspace images to obtain the analytical solution in \eqref{analytic}. It can be observed from the impulse response of $D_{w}$ in figure that its zeros lie exactly on the edge locations of phantom which verifies our assumption from [Greg's paper]. The reconstructed phantom is shown in figure.
%\begin{figure}
%\begin{subfigure}{0.15\textwidth}
%  \centering
%  \includegraphics[width=2.5cm,height=2.5cm]{atb_ph}
%  \caption{$A_{H}$b}
%  \label{fig:sfig1}
%\end{subfigure}%
%\begin{subfigure}{0.15\textwidth}
%  \centering
%  \includegraphics[width=2.5cm,height=2.5cm]{reconstruction}
%  \caption{Recon}
%  \label{fig:sfig2}
%\end{subfigure}
%\begin{subfigure}{0.15\textwidth}
 % \centering
% \includegraphics[width=2.5cm,height=2.5cm]{Phantom_edge}
 % \caption{Edge set}
%  \label{fig:sfig3}
%\end{subfigure}
%\caption{Phantom results}
%\label{fig:fig}
%\end{figure}
% \subsection{MR images}
Multi-channel brain MRI data was collected from five subjects at University of Iowa Hospitals using 3D T2 CUBE sequence with Cartesian readouts using a 12-channel head coil. The data from four subjects was used for training, while the data from the fifth subject was used for testing. We retrospectively undersampled the phase encodes to train and test the framework; we note that this approach is completely consistent with a future prospective acquisition, where a subset of phase encodes can be pre-selected and acquired. All the experiments were performed with variable-density Cartesian random sampling mask with different undersampling factors. We chose 30 most informative slices from each coil, yielding 360 from each subject to form a training dataset of 1440 slices.

%--------BEGIN FIGURE 2 --------------

\begin{figure}[t!]
\includegraphics[width=.5\textwidth]{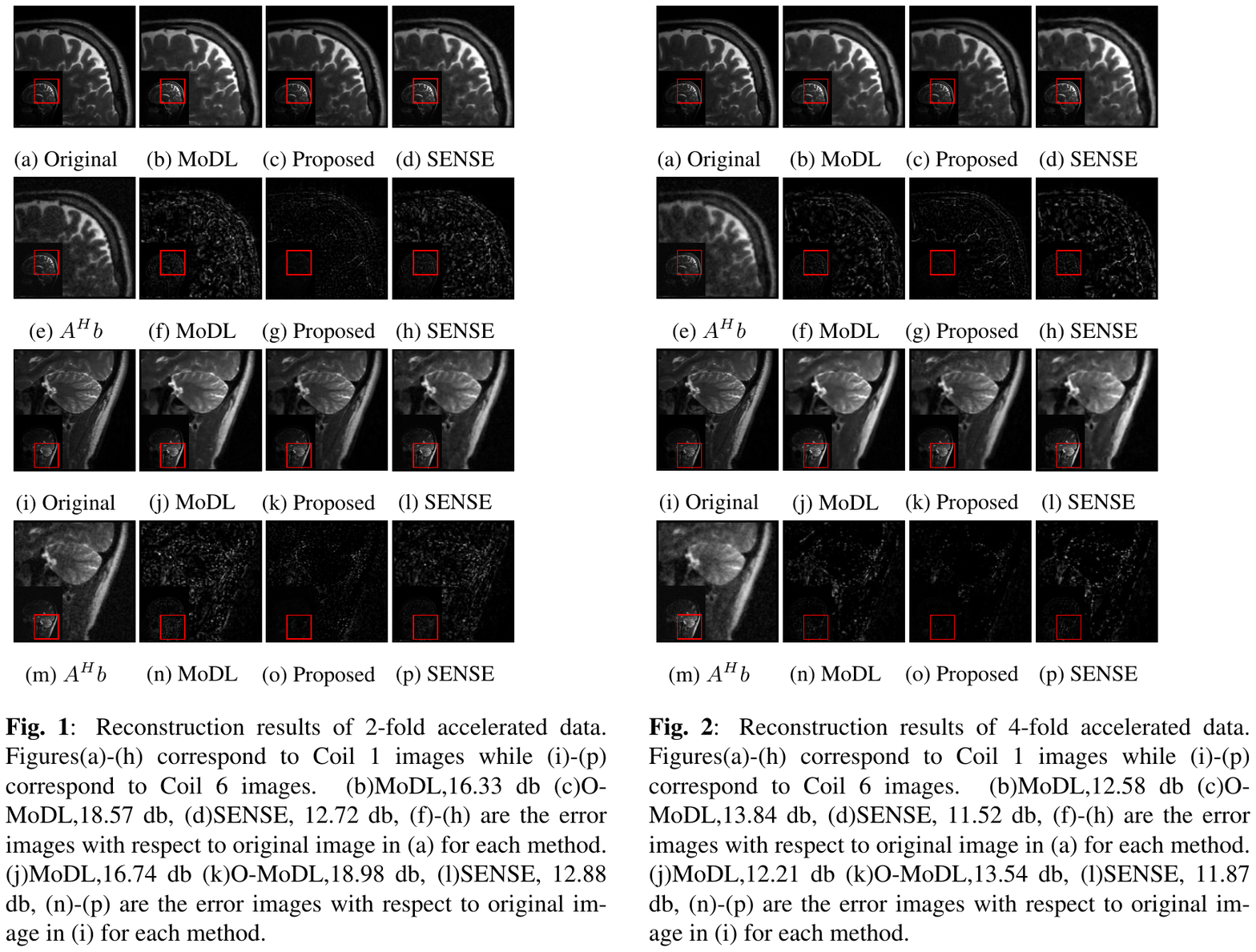}\vspace{-1.2em}
\caption{Reconstruction results of four-fold accelerated data. Figures(a)-(h) correspond to Coil 1 images while (i)-(p) correspond to Coil 6 images. (b) MoDL:12.58 db (c) O-MoDL:13.84 db, (d) SENSE: 11.52 db, (f)-(h) are the error images. (j) MoDL,12.21 db (k) O-MoDL,13.54 db, (l) SENSE, 11.87 db, (n)-(p) are the error images.} \vspace{-1.2em}
\end{figure}
%--------END FIGURE 2 --------------

%--------BEGIN TABLE-------------
\begin{table}[]
\fontsize{8}{16}
\selectfont
\centering
\begin{tabular}{cccc}
\hline
\multicolumn{4}{c}{2-fold/4-fold accelerated data} \\ \hline
%Recon time/image (sec) & 10 & 10 & 10 \\ \hline
Noise ($\sigma$) & MoDL & \textbf{Proposed} & SENSE   \\ \hline
10               &  16.42/12.46    &   \textbf{18.30/13.71}       & 11.7/11.44         \\ 
11               &  16.35/12.22    &   \textbf{18.29/13.68}      &  11.06/10.54       \\ 
12               &  15.92/11.91    &   \textbf{18.27/13.68}       & 10.66/9.89        \\ 
13               &  15.77/11.51    &   \textbf{18.25/13.66}       & 10.48/9.76        \\ \hline
\end{tabular}
\caption{Quantitative comparison of MoDL, Proposed, and SENSE reconstructions. The SNR is reported in dB and the two figures correspond to two fold and four fold acceleration.}\vspace{-1em}
\end{table}
%--------END TABLE---------------

We trained a network with an $\mathcal N_w$ block consisting of five convolution and five deconvolution layers as described in section \ref{sec:model}. Each layer consisted of 64 complex 3 x 3 filters. The iterative model was unrolled with five iterations, which was trained to minimize the mean square error between the output and the non-undersampled images. The training took 7 hours on an NVIDIA Tesla p100 GPU on 1440 slices. By contrast, the run time for reconstucting a single image is only 50 milliseconds. We compare the performance of O-MoDL against our image domain deep learning method (MoDL) and SENSE reconstructions for two-fold and four-fold accelerated data. For fair comparisons, we train both the networks with same number of iterations and trainable parameters. The SNR improvement in each case can be observed in table 1. Fig. 2 and 3 show reconstructions of different coil images from two and four fold accelerated data respectively. Results show that the proposed scheme can provide improved reconstructions over the MoDL scheme \cite{aggarwal2017modl}, which was trained exactly in the same fashion.

\vspace{-1em}
 \section{Conclusion}
We proposed an off-the-grid deep learning architecture for model based MR image reconstruction. Unlike many of the current CNN architectures, the algorithm works in the Fourier domain. The algorithm is a non-linear extension of recent structured low-rank off-the-grid methods, which rely on annihilation relations in the Fourier domain resulting from continuous domain image properties. Unlike the linear annihilation relations that are self learned in structured low-rank setting, the proposed framework learns non-linear annihilation relations in the Fourier domain from exemplar data; the non-linearities facilitate the the generalization of the annihilation properties to images unseen by the training algorithm, eliminating the need for self-learning the weights. The main benefit of the proposed scheme is the quite significant reduction in run time, compared to structured low-rank algorithms. The preliminary experimental comparisons demonstrate the improvements offered by the proposed scheme over image domain MoDL framework.             
\bibliographystyle{IEEEbib}

\bibliography{refs}

\begin{thebibliography}{1}

\bibitem{gregSIAM2016}
Greg Ongie and Mathews Jacob,
\newblock ``{Off-the-Grid Recovery of Piecewise Constant Images from Few
  Fourier Samples},''
\newblock {\em SIAM on Imag. Sci.}, vol. 9, no. 3, pp. 1004----1041, 2016.

\bibitem{haldar2014low}
Justin~P Haldar,
\newblock ``Low-rank modeling of local $ k $-space neighborhoods (loraks) for
  constrained mri,''
\newblock {\em IEEE transactions on medical imaging}, vol. 33, no. 3, pp.
  668--681, 2014.

\bibitem{ye2018deep}
Jong~Chul Ye, Yoseob Han, and Eunju Cha,
\newblock ``Deep convolutional framelets: A general deep learning framework for
  inverse problems,''
\newblock {\em SIAM Journal on Imaging Sciences}, vol. 11, no. 2, pp.
  991--1048, 2018.

\bibitem{unser2017}
Kyong~Hwan Jin, Michael~T. McCann, Emmanuel Froustey, and Michael Unser,
\newblock ``{Deep Convolutional Neural Network for Inverse Problems in
  Imaging},''
\newblock vol. 29, pp. 4509--4522, 2017.

\bibitem{aggarwal2017modl}
Hemant~Kumar Aggarwal, Merry~P Mani, and Mathews Jacob,
\newblock ``Modl: Model based deep learning architecture for inverse
  problems,''
\newblock {\em arXiv preprint arXiv:1712.02862}, 2017.

\bibitem{han2018k}
Yoseob Han and Jong~Chul Ye,
\newblock ``k-space deep learning for accelerated mri,''
\newblock {\em arXiv preprint arXiv:1805.03779}, 2018.

\end{thebibliography}

\end{document}